\pdfoutput=1 %

\documentclass[11pt,a4paper]{article}
\usepackage[hyperref]{acl2020}
\usepackage{times}
\usepackage{latexsym}

\usepackage{microtype}

\aclfinalcopy %

\usepackage{graphicx}
\usepackage{subcaption}
\usepackage{xspace}

\newcommand{\cls}{\texttt{[CLS]}\xspace}

\newcommand{\jointmulti}{{\small \textsc{JointMulti}}\xspace}
\newcommand{\jointpair}{{\small \textsc{JointPair}}\xspace}
\newcommand{\monotrans}{{\small \textsc{MonoTrans}}\xspace}
\newcommand{\clwe}{{\small \textsc{CLWE}}\xspace}

\newenvironment{enumeratesquish}{\begin{enumerate}{\Roman{enumi}}{\setlength{\itemsep}{-0.2em}\setlength{\labelwidth}{1.5em}\setlength{\leftmargin}{\labelwidth}\addtolength{\leftmargin}{\labelsep}}}{\end{enumerate}}

\usepackage[utf8]{inputenc}
\usepackage{CJKutf8}

\usepackage{multirow}
\usepackage{booktabs}

\title{On the Cross-lingual Transferability of Monolingual Representations}

\author{Mikel Artetxe$^{\dag}$\Thanks{ Work done as an intern at DeepMind.}~, \ Sebastian Ruder$^{\ddag}$~, \ Dani Yogatama$^{\ddag}$ \\
$^{\dag}$HiTZ Center, University of the Basque Country (UPV/EHU)\\
$^{\ddag}$DeepMind\\
\texttt{mikel.artetxe@ehu.eus} \\ 
\texttt{\{ruder,dyogatama\}@google.com}
}

\date{}

\begin{document}
\maketitle
\begin{abstract}
State-of-the-art unsupervised multilingual models 
(e.g., multilingual BERT)
have been shown to generalize in a zero-shot cross-lingual setting.
This generalization ability
has been attributed to the use of a shared subword vocabulary 
and joint training across multiple languages giving rise to deep multilingual abstractions.
We evaluate this hypothesis by designing an alternative 
approach that transfers a monolingual model to new languages at the lexical level.
More concretely, we first train a transformer-based masked language model
on one language, and transfer it to a new
language by learning a new embedding matrix
with the same masked language modeling objective---freezing
parameters of all other layers.
This approach does not rely on a shared vocabulary or joint training. 
However, we show that it is competitive with multilingual BERT 
on standard cross-lingual classification benchmarks 
and on a new Cross-lingual Question Answering Dataset (XQuAD). 
Our results contradict
common beliefs of the basis of the generalization ability
of multilingual models and
suggest that deep monolingual models learn 
some abstractions that generalize across languages.
We also release XQuAD as a more comprehensive cross-lingual benchmark, which comprises 240 paragraphs and 1190 
question-answer pairs from SQuAD v1.1 
translated into ten languages by professional translators.

\end{abstract}

\begin{figure*}[!htb]
    \begin{subfigure}{.22\linewidth}
      \centering
         \includegraphics[height=1.4in]{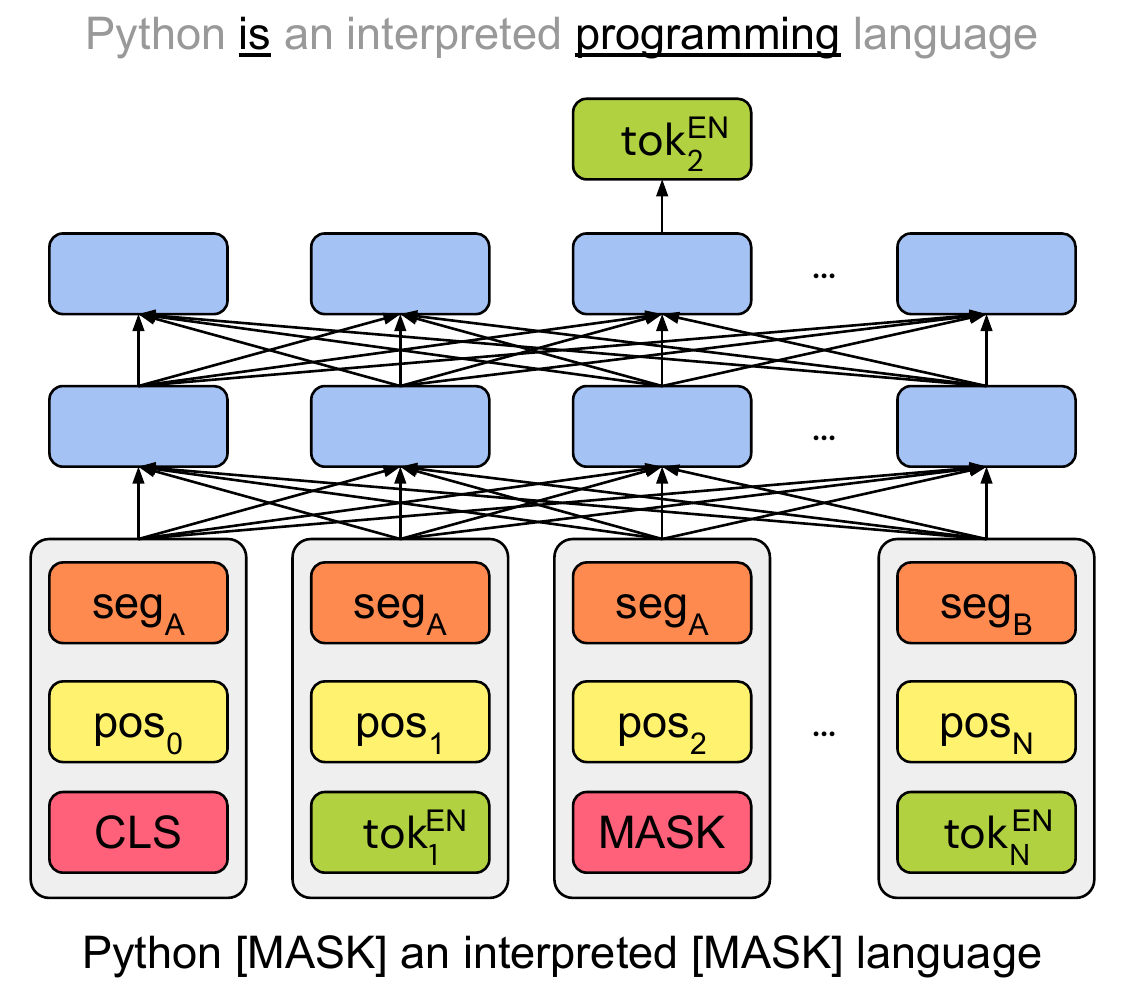}
    \caption{English pre-training} \label{fig:pre-training}
    \end{subfigure}
    \hspace*{0.3cm}
    \begin{subfigure}{.22\linewidth}
      \centering
         \includegraphics[height=1.4in]{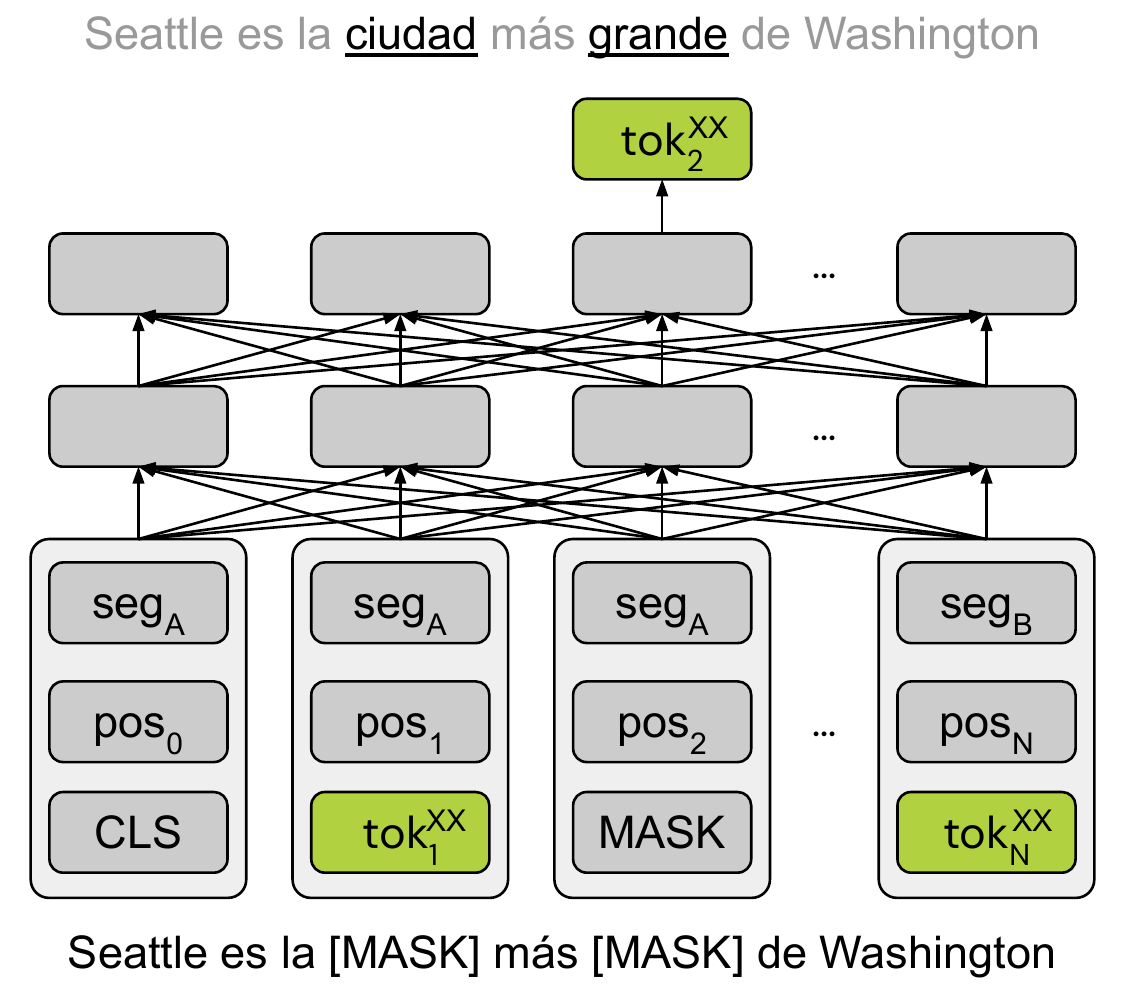}
    \caption{$L_2$ embedding learning} \label{fig:emb-learning}
    \end{subfigure}
    \hspace*{0.25cm}
    \begin{subfigure}{.22\linewidth}
      \centering
         \includegraphics[height=1.4in]{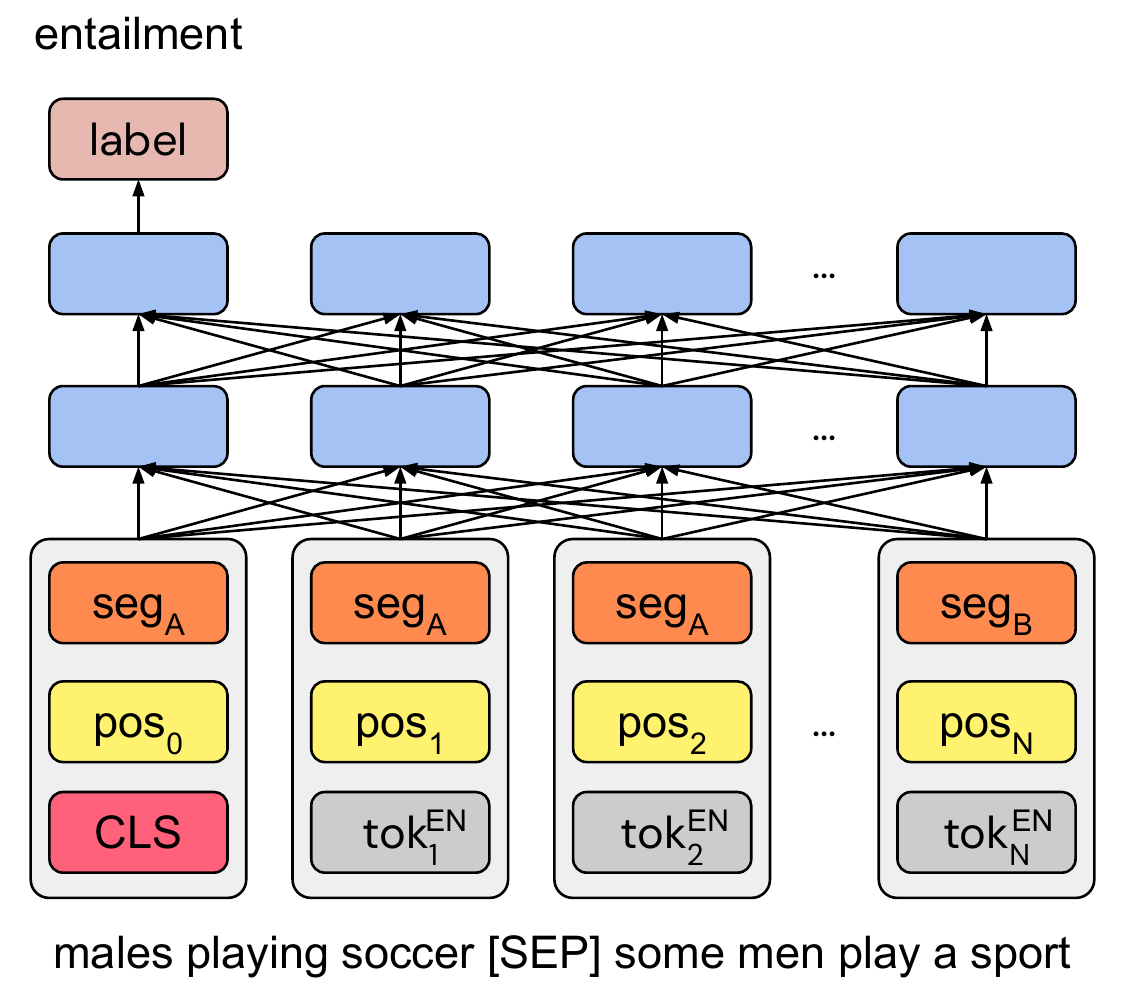}
    \caption{English fine-tuning} \label{fig:fine-tuning}
    \end{subfigure}
    \hspace*{0.32cm}
    \begin{subfigure}{.22\linewidth}
      \centering
         \includegraphics[height=1.4in]{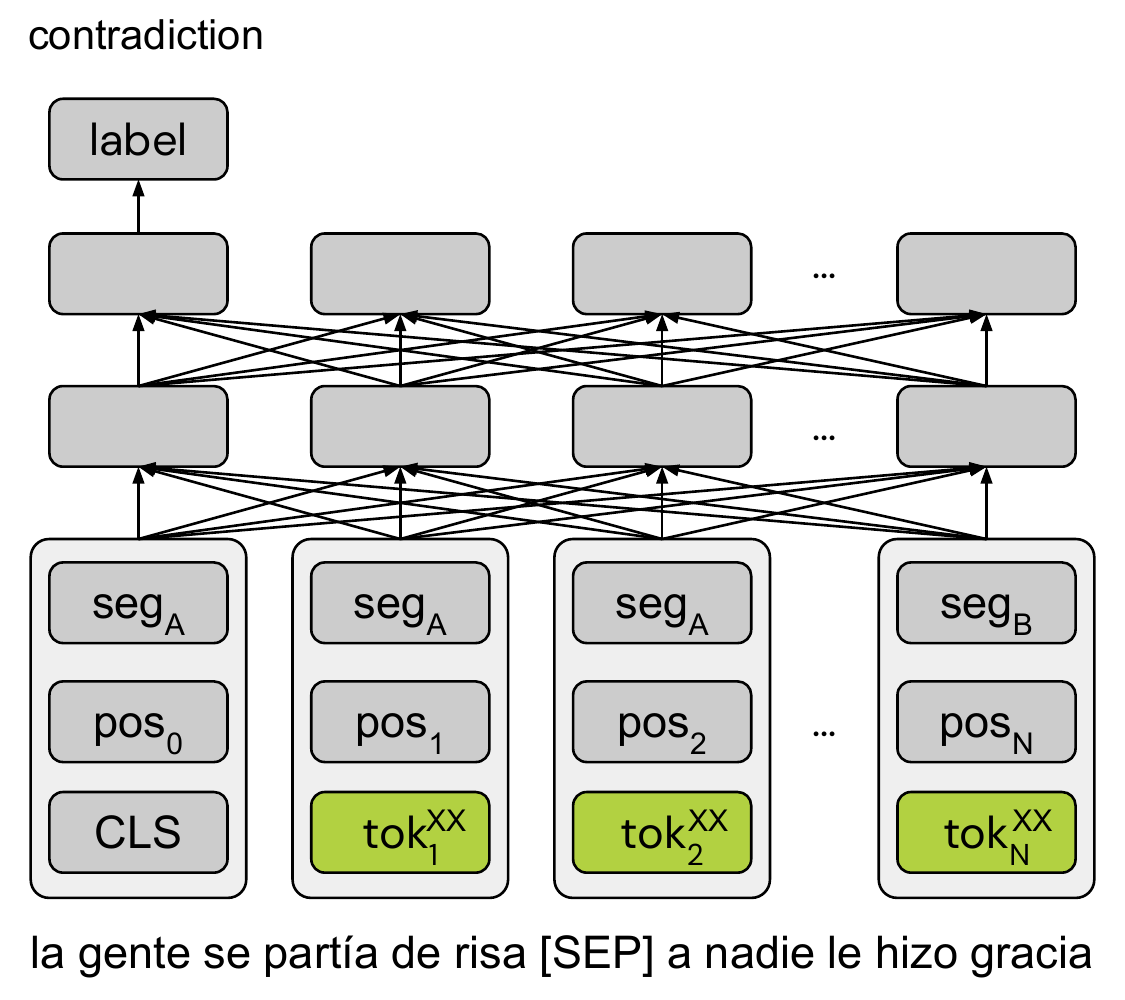}
    \caption{Zero-shot transfer to $L_2$} \label{fig:zero-shot-transfer}
    \end{subfigure}
    \caption{Four steps for zero-shot cross-lingual transfer: (i) pre-train a monolingual transformer model in English akin to BERT; (ii) freeze the transformer body and learn new token embeddings from scratch for a second language using the same training objective over its monolingual corpus; (iii) fine-tune the model on English while keeping the embeddings frozen; and (iv) zero-shot transfer it to the new language by swapping the token embeddings.}
\label{fig:overview}
\end{figure*}

\section{Introduction}

Multilingual pre-training methods such as multilingual BERT (mBERT, \citealp{devlin2018bert}) have been successfully used for zero-shot cross-lingual transfer \citep{Pires2019,lample2019cross}. These methods work by jointly training a transformer model \citep{vaswani2017attention} to perform masked language modeling (MLM) in multiple languages, which is then fine-tuned on a downstream task using labeled data in a single language---typically English. As a result of the multilingual pre-training, the model is able to generalize to other languages, even if it has never seen labeled data in those languages. 
Such a cross-lingual generalization 
ability is surprising, as there is no explicit cross-lingual 
term in the underlying training objective. %
In relation to this, \citet{Pires2019} hypothesized that:
\vspace{1mm}
\begin{center}
{\small
\begin{minipage}{2.5in}{\em \ldots having word pieces used in all languages (numbers, URLs, etc), which have to be mapped to a shared space forces the co-occurring pieces to also be mapped to a shared space, thus spreading the effect to other word pieces, until different languages are close to a shared space. \\
\ldots mBERT's ability to generalize cannot be attributed solely to vocabulary memorization, and that it must be learning a deeper multilingual representation.}\end{minipage}}
\end{center}
\citet{iclr2020_multilingual} echoed this sentiment, and
\citet{Wu2019} further observed that 
mBERT performs better in languages that share many subwords. %
As such, the current consensus of the cross-lingual
generalization ability of mBERT
is based on a combination of three factors: 
(i) shared vocabulary items that act as anchor points; 
(ii) joint training across multiple languages that spreads this effect; which ultimately yields (iii) 
deep cross-lingual representations that generalize 
across languages and tasks.

In this paper, we empirically test this hypothesis 
by designing an alternative approach that 
violates all of these assumptions.
As illustrated in Figure \ref{fig:overview},
our method starts with a monolingual transformer 
trained with MLM, which we transfer to a new language by learning a new embedding matrix through MLM in the new language 
while freezing parameters of all other layers.
This approach only learns
new lexical parameters and does not rely 
on shared vocabulary items nor joint learning. However, 
we show that it 
is competitive with joint 
multilingual pre-training across 
standard zero-shot cross-lingual transfer 
benchmarks (XNLI, MLDoc, and PAWS-X).

We also experiment with a new Cross-lingual Question Answering Dataset (XQuAD), which
consists of 240 paragraphs and
1190 question-answer pairs from SQuAD v1.1 \citep{squad} 
translated into ten languages by professional translators. 
Question answering as a task
is a classic probe for language understanding.
It has also been found to be less susceptible to annotation artifacts commonly found in other benchmarks \citep{Kaushik2018,Gururangan2018}. 
We believe that
XQuAD can serve as a more comprehensive cross-lingual benchmark and make it publicly available at \url{https://github.com/deepmind/xquad}.
Our results on XQuAD show that the monolingual transfer
approach can be made competitive with mBERT by learning second language-specific transformations via adapter modules \citep{Rebuffi2017}.

Our contributions in this paper are as follows:
(i) we propose a method to transfer monolingual representations to new languages in an unsupervised fashion (\S{\ref{sec:proposed}})\footnote{This is particularly useful for low-resource languages,
since many pre-trained models are currently in English.};
(ii) we show
that neither a shared subword vocabulary nor joint multilingual training
is necessary for zero-shot transfer 
and find that the effective vocabulary size per language 
is an important factor for learning multilingual models
(\S{\ref{sec:experiments}} and \S{\ref{sec:xquad}});
(iii) we show that monolingual models learn abstractions 
that generalize across languages (\S{\ref{sec:discussion}});
and (iv) we present a new cross-lingual question answering dataset (\S\ref{sec:xquad}).

\section{Cross-lingual Transfer of Monolingual Representations} \label{sec:proposed}

In this section, we propose an approach to transfer a pre-trained monolingual model 
in one language $L_1$ (for which both task supervision and a monolingual corpus are available) to a second language $L_2$ (for which only a monolingual corpus is available).
The method serves as a counterpoint to existing joint multilingual models, as it works by aligning new lexical parameters
to a monolingually trained deep model.

As illustrated in Figure \ref{fig:overview}, our proposed method consists of four steps: 
\begin{enumeratesquish}
\item Pre-train a monolingual BERT (i.e. a transformer) in $L_1$ with masked language modeling (MLM) and next sentence prediction (NSP) objectives on an unlabeled $L_1$ corpus.
\item Transfer the model to a new language by learning new token embeddings \emph{while freezing the transformer body} with the same training objectives (MLM and NSP)
on an unlabeled $L_2$ corpus.
\item Fine-tune the transformer for a downstream task using labeled data in $L_1$, \emph{while keeping the $L_1$ token embeddings frozen}.
\item Zero-shot transfer the resulting model to $L_2$ by swapping the $L_1$ token embeddings with the $L_2$ embeddings learned in Step 2.
\end{enumeratesquish}

We note that, unlike mBERT, we use a separate subword vocabulary for each language, which is trained on its respective monolingual corpus, so the model has no notion of shared subwords. %
However, the special \cls, \texttt{[SEP]}, \texttt{[MASK]}, \texttt{[PAD]}, and \texttt{[UNK]} symbols are shared across languages, and fine-tuned in Step 3.\footnote{The rationale behind this is that special symbols are generally task dependent, and given that the fine-tuning in downstream tasks is done exclusively in English, we need to share these symbols to zero-shot transfer to other languages.} %
We observe further improvements on several downstream tasks
using the following extensions to the above method.

\paragraph{Language-specific position embeddings.}
The basic approach does not take into account 
different word orders commonly found in different languages,
as it reuses the position embeddings in $L_1$ for $L_2$. 
We relax this restriction by
learning a separate set of position embeddings for $L_2$ in Step 2 (along with $L_2$ token embeddings).\footnote{We also freeze the $L_1$ position embeddings in Step 3 accordingly, and the $L_2$ position 
embeddings are plugged in together with the token embeddings in Step 4.} We treat the \cls symbol
as a special case. In the original implementation, 
BERT treats \cls as a regular word with its 
own position and segment embeddings, even if it 
always appears in the first position.
However, this does not provide any extra capacity to the model, as the same position and segment embeddings are always added up to the \cls embedding.
Following this observation,
we do
not use any position and 
segment embeddings for the \cls symbol.

\paragraph{Noised fine-tuning.}
The transformer body in our proposed method 
is only trained with $L_1$ embeddings as its input layer, 
but is used with $L_2$ embeddings at test time. To make the model more robust to this mismatch, 
we add Gaussian noises 
sampled from the standard normal distribution
to the word, position, and segment embeddings 
\emph{during the fine-tuning step} (Step 3). 

\paragraph{Adapters.}
We also investigate the possibility of allowing the model
to learn
better deep representations of $L_2$, 
while retaining the alignment with $L_1$ using residual adapters
\citep{Rebuffi2017}. 
Adapters are small task-specific bottleneck layers that are added between layers of a pre-trained model. 
During fine-tuning, the original model parameters 
are frozen, and only parameters of the adapter modules
are learned.
In Step 2, when we transfer the $L_1$ transformer
to $L_2$, we add a feed-forward adapter module after the 
projection following multi-headed attention 
and after the two feed-forward layers 
in each transformer layer, similar to \citet{Houlsby2019}.
Note that the original transformer body is still frozen,
and only parameters of the adapter modules are trainable (in addition 
to the embedding matrix in $L_2$).

\section{Experiments}
\label{sec:experiments}
Our goal is to  
evaluate the performance of different multilingual 
models in the zero-shot cross-lingual
setting to better understand the source
of their generalization ability.
We describe the models that we compare (\S\ref{sec:expmodels}),
the experimental setting (\S\ref{sec:expsetup}),
and the results on three classification datasets: XNLI (\S{\ref{subsec:xnli}}), MLDoc (\S{\ref{subsec:mldoc}}) and PAWS-X (\S{\ref{subsec:pawsx}}). 
We discuss experiments on our new XQuAD dataset 
in \S{\ref{sec:xquad}}.
In all experiments, we fine-tune a pre-trained model 
using labeled training examples in English, and evaluate on
test examples in other languages via zero-shot transfer.

\subsection{Models}
\label{sec:expmodels}
We compare four main models in our experiments:
\paragraph{Joint multilingual models (\jointmulti).} A multilingual BERT model trained jointly on 15 languages\footnote{We use all languages that
are included in XNLI \cite{conneau2018xnli}.}. This model is analogous to mBERT and
closely related to other variants like XLM.

\paragraph{Joint pairwise bilingual models (\jointpair).} A multilingual BERT model trained jointly on two languages (English and another language). This serves to control the effect of having multiple languages in joint training. 
At the same time, it provides a joint system that is directly comparable to the monolingual transfer approach in \S\ref{sec:proposed}, which also operates on two languages.

\paragraph{Cross-lingual word embedding mappings (\clwe).}
The method we described in \S\ref{sec:proposed} operates at the lexical level, 
and can be seen as a form of learning cross-lingual word embeddings 
that are aligned to a monolingual transformer body. 
In contrast to this approach, standard cross-lingual word embedding mappings 
first align monolingual lexical spaces 
and then learn a multilingual deep model on top of this space.
We also include a method based on this alternative approach where we 
train skip-gram embeddings for each language, and map them to a shared space using VecMap \citep{artetxe2018robust}.\footnote{We use the \textit{orthogonal} mode in VecMap and map all languages into English.}
We then train an English BERT model using MLM and NSP 
on top of the frozen mapped embeddings.
The model is then fine-tuned 
using English labeled data while 
keeping the embeddings frozen. We 
zero-shot transfer to a new language by 
plugging in its respective mapped embeddings.

\paragraph{Cross-lingual transfer of monolingual models (\monotrans).} Our method described in \S\ref{sec:proposed}. We use English as $L_1$ and try multiple variants with different extensions.

\begin{table*}[t]
\begin{center}
\begin{small}
\addtolength{\tabcolsep}{-3.5pt}
  \begin{tabular}{rcrlrcrccccccccccccccrcr}
    \toprule
    & & & & & en & & fr & es & de & el & bg & ru & tr & ar & vi & th & zh & hi & sw & ur & & avg & \\
    \midrule
    & \multirow{2}{*}{\shortstack{Prev\\work}}
    && mBERT && 81.4 & & - & 74.3 & 70.5 & - & - & - & - & 62.1 & - & - & 63.8 & - & - & 58.3 && - & \\
    &&& XLM (MLM) && \underline{83.2} & & \underline{76.5} & 76.3 & 74.2 & 73.1 & \underline{74.0} & \underline{73.1} & 67.8 & 68.5 & 71.2 & \underline{69.2} & 71.9 & 65.7 & \underline{64.6} & \underline{63.4} && \underline{71.5} & \\
    \midrule
    & \multirow{4}{*}{\shortstack{\clwe}}
    && 300d ident && 82.1 && 67.6 & 69.0 & 65.0 & 60.9 & 59.1 & 59.5 & 51.2 & 55.3 & 46.6 & 54.0 & 58.5 & 48.4 & 35.3 & 43.0 && 57.0 & \\
    &&& 300d unsup && 82.1 && 67.4 & 69.3 & 64.5 & 60.2 & 58.4 & 59.2 & 51.5 & 56.2 & 36.4 & 54.7 & 57.7 & 48.2 & 36.2 & 33.8 && 55.7 & \\
    &&& 768d ident && \textbf{82.4} && \textbf{70.7} & 71.1 & \textbf{67.6} & \textbf{64.2} & 61.4 & \textbf{63.3} & \textbf{55.0} & \textbf{58.6} & \textbf{50.7} & \textbf{58.0} & \textbf{60.2} & 54.8 & 34.8 & \textbf{48.1} && \textbf{60.1} & \\
    &&& 768d unsup && \textbf{82.4} && 70.4 & \textbf{71.2} & 67.4 & 63.9 & \textbf{62.8} & \textbf{63.3} & 54.8 & 58.3 & 49.1 & 57.2 & 55.7 & \textbf{54.9} & \textbf{35.0} & 33.9 && 58.7 & \\
    \midrule
    & \multirow{4}{*}{\shortstack{{\small \textsc{Joint}}\\{\small \textsc{Multi}}}}
    && 32k voc && 79.0 & & 71.5 & 72.2 & 68.5 & 66.7 & 66.9 & 66.5 & 58.4 & 64.4 & 66.0 & 62.3 & 66.4 & 59.1 & 50.4 & 56.9 && 65.0 & \\
    &&& 64k voc && 80.7 & & 72.8 & 73.0 & 69.8 & 69.6 & 69.5 & 68.8 & 63.6 & 66.1 & 67.2 & 64.7 & 66.7 & 63.2 & 52.0 & 59.0 && 67.1 & \\
    &&& 100k voc && 81.2 & & 74.5 & 74.4 & 72.0 & 72.3 & 71.2 & 70.0 & 65.1 & 69.7 & 68.9 & 66.4 & 68.0 & 64.2 & 55.6 & 62.2 && 69.0 & \\
    &&& 200k voc && \textbf{82.2} & & \textbf{75.8} & \textbf{75.7} & \textbf{73.4} & \textbf{74.0} & \textbf{73.1} & \textbf{71.8} & \textbf{67.3} & \textbf{69.8} & \textbf{69.8} & \textbf{67.7} & \textbf{67.8} & \textbf{65.8} & \textbf{60.9} & \textbf{62.3} && \textbf{70.5} & \\
    \midrule
    & \multirow{2}{*}{\shortstack{{\small \textsc{Joint}}\\{\small \textsc{Pair}}}}
    && Joint voc && 82.2 & & 74.8 & 76.4 & 73.1 & 72.0 & 71.8 & 70.2 & 67.9 & 68.5 & \underline{\textbf{71.4}} & \textbf{67.7} & 70.8 & 64.5 & \textbf{64.2} & \textbf{60.6} && 70.4 & \\
    &&& Disjoint voc && \textbf{83.0} & & \textbf{76.2} & \underline{\textbf{77.1}} & \underline{\textbf{74.4}} & \underline{\textbf{74.4}} & \textbf{73.7} & \textbf{72.1} & \textbf{68.8} & \underline{\textbf{71.3}} & 70.9 & 66.2 & \underline{\textbf{72.5}} & \underline{\textbf{66.0}} & 62.3 & 58.0 && \textbf{71.1} & \\
    \midrule
    & \multirow{4}{*}{\shortstack{{\small \textsc{Mono}}\\{\small \textsc{Trans}}}}
    && Token emb && 83.1 && 73.3 & 73.9 & 71.0 & 70.3 & 71.5 & 66.7 & 64.5 & 66.6 & 68.2 & 63.9 & 66.9 & 61.3 & 58.1 & 57.3 && 67.8 & \\
    &&& \quad + pos emb && \textbf{83.8} && 74.3 & 75.1 & 71.7 & 72.6 & 72.8 & 68.8 & 66.0 & 68.6 & \textbf{69.8} & 65.7 & 69.7 & 61.1 & 58.8 & 58.3 && 69.1 & \\
    &&& \quad + noising && 81.7 && 74.1 & 75.2 & 72.6 & \textbf{72.9} & 73.1 & 70.2 & 68.1 & 70.2 & 69.1 & \textbf{67.7} & \textbf{70.6} & 62.5 & \textbf{62.5} & \textbf{60.2} && \textbf{70.0} & \\
    &&& \quad + adapters && 81.7 && \textbf{74.7} & \textbf{75.4} & \textbf{73.0} & 72.0 & \textbf{73.7} & \textbf{70.4} & \underline{\textbf{69.9}} & \textbf{70.6} & 69.5 & 65.1 & 70.3 & \textbf{65.2} & 59.6 & 51.7 && 69.5 & \\
    \bottomrule
  \end{tabular}
\end{small}
\end{center}
\caption{XNLI results (accuracy). mBERT results are taken from the official BERT repository, while XLM results are taken from \citet{lample2019cross}. We bold the best result in each section and underline the overall best.}
\label{tab:results_xnli}
\end{table*}

\subsection{Setting}
\label{sec:expsetup}
\paragraph{Vocabulary.} We perform subword tokenization using the unigram model in SentencePiece \citep{kudo2018sentencepiece}.
In order to understand the 
effect of sharing subwords across languages and the size of the vocabulary, we train 
each model with various settings. 
We train 4 different \jointmulti models 
with a vocabulary of 32k, 64k, 100k, and 200k subwords. 
For \jointpair, we train one model with a joint vocabulary of 32k subwords, learned separately for each language pair, and another one with a disjoint vocabulary of 32k subwords per language, learned on its respective monolingual corpus.
The latter is directly comparable 
to \monotrans
in terms of vocabulary, 
in that it is restricted to two languages 
and uses the exact same disjoint vocabulary 
with 32k subwords per language. 
For \clwe, we use the same subword vocabulary 
and investigate two choices: (i) the number of 
embedding dimensions---300d (the standard in 
the cross-lingual embedding literature) and 768d (equivalent to the rest of the models); and
(ii) the self-learning initialization---weakly supervised 
(based on identically spelled words, \citealp{Sogaard2018}) and 
unsupervised (based on the intralingual similarity distribution, \citealp{artetxe2018robust}).

\paragraph{Pre-training data.} We use Wikipedia as our training corpus, 
similar to mBERT and XLM \citep{lample2019cross},
which we extract using the WikiExtractor tool.\footnote{\url{https://github.com/attardi/wikiextractor}} 
We do not perform any lowercasing or normalization.
When working with languages of different corpus sizes, we use the same upsampling strategy as \citet{lample2019cross} 
for both the subword vocabulary learning and the pre-training.

\paragraph{Training details.}
Our implementation is based on the BERT code from \citet{devlin2018bert}. For adapters, we build on the code by \citet{Houlsby2019}.
We use the model architecture of BERT$_{\textsc{base}}$,
similar to mBERT. %
We use the 
LAMB optimizer \cite{you2019reducing} and 
train on 64 TPUv3 chips for 250,000 steps using 
the same hyperparameters as \citet{you2019reducing}. 
We describe other training details in Appendix~\ref{app:hyperparameters}.
Our hyperparameter configuration is based on preliminary experiments 
on the development set of the XNLI dataset. 
We do not perform any exhaustive hyperparameter search, 
and use the exact same settings for all model variants, languages, and tasks.

\paragraph{Evaluation setting.}
We perform a single training and evaluation run for each model, 
and report results in the corresponding test set 
for each downstream task.
For \monotrans,
we observe stability issues 
when learning language-specific position embeddings 
for Greek, Thai and Swahili. The second step would occasionally fail to converge to a good solution. 
For these three languages, we run Step 2 of our proposed method  (\S\ref{sec:proposed}) three times and
pick the best model on the XNLI development set.

\subsection{XNLI: Natural Language Inference} \label{subsec:xnli}
In natural language inference (NLI), given two sentences (a premise and a hypothesis), 
the goal is to decide whether there is an \textit{entailment}, \textit{contradiction}, or \textit{neutral} 
relationship between them \citep{bowman2015large}.
We train all models on the MultiNLI 
dataset \cite{williams2018broad} in English and evaluate on XNLI \cite{conneau2018xnli}---a cross-lingual NLI 
dataset consisting of 2,500 development and 5,000 
test instances translated from English into 14 languages.

We report our results on XNLI in Table~\ref{tab:results_xnli} together with the previous results from mBERT and XLM.\footnote{mBERT covers 102 languages and has a shared vocabulary of 110k subwords. XLM covers 15 languages and uses a larger model size with a shared vocabulary of 95k subwords, which contributes to its better performance.}
We summarize our main findings below.
\paragraph{\jointmulti is comparable with the literature.} Our best \jointmulti model 
is substantially better than mBERT, and only one 
point worse (on average) than the unsupervised XLM model, 
which is larger in size.

\paragraph{A larger vocabulary is beneficial.} \jointmulti variants with a larger vocabulary perform better. %

\paragraph{More languages do not improve performance.} \jointpair models with a joint 
vocabulary perform comparably with \jointmulti. %

\paragraph{A shared subword vocabulary is \emph{not} necessary for joint multilingual pre-training.} The equivalent \jointpair models with 
a disjoint vocabulary for each language perform better.

\paragraph{\clwe performs poorly.}
Even if it is competitive in English, 
it does not transfer as well to other languages.
Larger dimensionalities and weak supervision improve \clwe,
but its performance is still below other models.

\paragraph{\monotrans is competitive with joint learning.} The basic version of \monotrans
is 3.3 points worse on average than its equivalent \jointpair model.
Language-specific position embeddings 
and noised fine-tuning reduce the gap to only 1.1 points.
Adapters mostly improve performance, except for low-resource languages 
such as Urdu, Swahili, Thai, and Greek.
In subsequent experiments, we include
results for all variants of \monotrans and \jointpair,
the best \clwe variant (768d ident), and \jointmulti with 32k and 200k voc.

\begin{table*}[t]
\begin{center}
\begin{small}
\addtolength{\tabcolsep}{-3.5pt}
  \begin{tabular}{rcrlrcrcccccrcrrcrccccrcr}
    \toprule
    &&&&& \multicolumn{9}{c}{MLDoc} &&& \multicolumn{8}{c}{PAWS-X} & \\
    \cmidrule{6-14} \cmidrule{17-24}
    & & & & & en & & fr & es & de & ru & zh && avg &&& en && fr & es & de & zh && avg & \\
    \midrule
    & Prev work && mBERT && - && 83.0 & 75.0 & 82.4 & 71.6 & 66.2 && - &&& 93.5 && 85.2 & 86.0 & 82.2 & 75.8 && 84.5 & \\
    \midrule
    & \multirow{1}{*}{\shortstack{\clwe}}
    && 768d ident && \underline{94.7} && \underline{87.3} & \underline{77.0} & 88.7 & 67.6 & 78.3 && \underline{82.3} &&& 92.8 && 85.2 & 85.5 & 81.6 & 72.5 && 83.5 & \\
    \midrule
    & \multirow{2}{*}{\shortstack{{\small \textsc{Joint}}\\{\small \textsc{Multi}}}}
    && 32k voc && \textbf{92.6} && 81.7 & 75.8 & 85.4 & 71.5 & \textbf{66.6} && 78.9 &&& 91.9 && 83.8 & 83.3 & 82.6 & 75.8 && 83.5 & \\
    &&& 200k voc && 91.9 && \textbf{82.1} & \textbf{80.9} & \underline{\textbf{89.3}} & \underline{\textbf{71.8}} & 66.2 && \textbf{80.4} &&& \textbf{93.8} && \textbf{87.7} & \textbf{87.5} & \textbf{87.3} & \textbf{78.8} && \textbf{87.0} & \\
    \midrule
    & \multirow{2}{*}{\shortstack{{\small \textsc{Joint}}\\{\small \textsc{Pair}}}}
    && Joint voc && 93.1 && 81.3 & 74.7 & \textbf{87.7} & \textbf{71.5} & \textbf{80.7} && \textbf{81.5} &&& 93.3 && 86.1 & 87.2 & 86.0 & \underline{\textbf{79.9}} && 86.5 & \\
    &&& Disjoint voc && \textbf{93.5} && \textbf{83.1} & \textbf{78.0} & 86.6 & 65.5 & 78.1 && 80.8 &&& \textbf{94.0} && \underline{\textbf{88.4}} & \underline{\textbf{88.6}} & \underline{\textbf{87.5}} & 79.3 && \underline{\textbf{87.5}} & \\
    \midrule
    & \multirow{4}{*}{\shortstack{{\small \textsc{Mono}}\\{\small \textsc{Trans}}}}
    && Token emb && 93.5 && \textbf{84.0} & \textbf{76.9} & 88.7 & 60.6 & \underline{\textbf{83.6}} && \textbf{81.2} &&& 93.6 && 87.0 & 87.1 & 84.2 & 78.2 && 86.0 & \\
    &&& \quad + pos emb && \textbf{93.6} && 79.7 & 75.7 & 86.6 & 61.6 & 83.0 && 80.0 &&& \underline{\textbf{94.3}} && \textbf{87.3} & \textbf{87.6} & \textbf{86.3} & \textbf{79.0} && \textbf{86.9} & \\
    &&& \quad + noising && 88.2 && 81.3 & 72.2 & 89.4 & \textbf{63.9} & 65.1 && 76.7 &&& 88.0 && 83.3 & 83.2 & 81.8 & 77.5 && 82.7 & \\
    &&& \quad + adapters && 88.2 && 81.4 & 76.4 & \textbf{89.6} & 63.1 & 77.3 && 79.3 &&& 88.0 && 84.1 & 83.0 & 81.5 & 73.5 && 82.0 & \\
    \bottomrule
  \end{tabular}
\end{small}
\end{center}
\caption{MLDoc and PAWS-X results (accuracy). mBERT results are from \citet{Eisenschlos2019} for MLDoc and from \citet{yang2019pawsx} for PAWS-X, respectively. We bold the best result in each section with more than two models and underline the overall best result.}
\label{tab:results_mldoc_paws}
\end{table*}

\subsection{MLDoc: Document Classification} \label{subsec:mldoc}

In MLDoc \citep{schwenk2018corpus}, the task is to classify
documents into one of four different 
genres: \textit{corporate/industrial}, \textit{economics}, \textit{government/social}, and \textit{markets}.
The dataset is an improved version of the Reuters benchmark \citep{klementiev2012inducing}, and
consists of 1,000 training and 4,000 test documents in 7 languages. 

We show the results of our MLDoc experiments in~Table~\ref{tab:results_mldoc_paws}.
In this task, we observe that 
simpler models tend to perform better, and the best overall results are from \clwe. 
We believe that this can be attributed 
to: (i) the superficial nature of the task itself, 
as a model can rely on a few keywords 
to identify the genre of an input document
without requiring any high-level understanding
and (ii) the small size of the training set. 
Nonetheless, all of the four model 
families obtain generally similar results, 
corroborating our previous findings that
joint multilingual pre-training and 
a shared vocabulary are not needed to achieve good performance.

\subsection{PAWS-X: Paraphrase Identification} \label{subsec:pawsx}

PAWS is a dataset that contains pairs of sentences with a high lexical overlap \citep{zhang2019paws}.
The task is to predict whether each pair is
a paraphrase or not.
While the original dataset 
is only in English, PAWS-X \citep{yang2019pawsx} 
provides human translations into six languages.

We evaluate our models on this dataset and show 
our results in Table \ref{tab:results_mldoc_paws}.
Similar to experiments on other datasets, 
\monotrans is competitive with the best joint variant, 
with a difference of only 0.6 points 
when we learn language-specific position embeddings.

\section{XQuAD: Cross-lingual Question Answering Dataset} \label{sec:xquad}

Our classification experiments demonstrate that
\monotrans is competitive with \jointmulti and \jointpair, despite being multilingual at the embedding layer only (i.e. the transformer body is trained exclusively on English). 
One possible explanation for this behaviour is that existing cross-lingual benchmarks are flawed and solvable at the lexical level. For example, previous work has shown that models trained on MultiNLI---from which XNLI was derived---learn to exploit superficial cues in the data \cite{Gururangan2018}.

To better understand the cross-lingual generalization ability 
of these models, 
we create a new Cross-lingual Question Answering Dataset (XQuAD). Question answering is a classic probe for natural language understanding \cite{Hermann2015} and has been shown to be less susceptible to annotation artifacts than other popular tasks \cite{Kaushik2018}. 
In contrast to existing classification benchmarks, extractive question
answering requires identifying relevant answer spans 
in longer context paragraphs, thus requiring some degree of structural transfer across languages.

XQuAD consists of a subset of 240 paragraphs and 1190 question-answer pairs 
from the development set of SQuAD v1.1\footnote{We choose 
SQuAD 1.1 to avoid translating unanswerable questions.} together with their translations into 
ten languages: Spanish, German, Greek, Russian, Turkish, Arabic, Vietnamese, Thai, Chinese, and Hindi.
Both the context paragraphs and the questions are translated by professional human translators from Gengo\footnote{\url{https://gengo.com}}.
In order to facilitate easy annotations of answer spans,
we choose the most frequent answer for each question and mark its beginning and end in the context paragraph using special symbols, instructing translators to keep these symbols in the relevant positions in their translations.
Appendix~\ref{app:xquad_dataset}
discusses the dataset in more details.

\begin{table*}[t]
\begin{center}
\begin{small}
\addtolength{\tabcolsep}{-3.5pt}
  \begin{tabular}{rcrlrcrccccccccccrcr}
    \toprule
    & & & & & en && es & de & el & ru & tr & ar & vi & th & zh & hi && avg & \\
    \midrule
    & && mBERT && \underline{88.9} && \underline{75.5} & 70.6 & 62.6 & 71.3 & 55.4 & 61.5 & \underline{69.5} & 42.7 & 58.0 & 59.2 && 65.0 \\
    \midrule
    
    & \multirow{1}{*}{\shortstack{\clwe}}
    && 768d ident && 84.2 && 58.0 & 51.2 & 41.1 & 48.3 & 24.2 & 32.8 & 29.7 & 23.8 & 19.9 & 21.7 && 39.5 & \\
    \midrule
    & \multirow{2}{*}{\shortstack{{\small \textsc{Joint}}\\{\small \textsc{Multi}}}}
    && 32k voc && 79.3 && 59.5 & 60.3 & 49.6 & 59.7 & 42.9 & 52.3 & 53.6 & 49.3 & 50.2 & 42.3 && 54.5 & \\
    &&& 200k voc && \textbf{82.7} && \textbf{74.3} & \textbf{71.3} & \textbf{67.1} & \textbf{70.2} & \textbf{56.6} & \textbf{64.8} & \textbf{67.6} & \underline{\textbf{58.6}} & \textbf{51.5} & \textbf{58.3} && \textbf{65.7} & \\
    \midrule
    & \multirow{2}{*}{\shortstack{{\small \textsc{Joint}}\\{\small \textsc{Pair}}}}
    && Joint voc && 82.8 && 68.3 & \underline{\textbf{73.6}} & 58.8 & 69.8 & 53.8 & 65.3 & \underline{\textbf{69.5}} & \textbf{56.3} & 58.8 & \textbf{57.4} && 64.9 & \\
    &&& Disjoint voc && \textbf{83.3} && \textbf{72.5} & 72.8 & \textbf{67.3} & \underline{\textbf{71.7}} & \textbf{60.5} & \underline{\textbf{66.5}} & 68.9 & 56.1 & \textbf{60.4} & 56.7 && \underline{\textbf{67.0}} & \\
    \midrule
    & \multirow{4}{*}{\shortstack{{\small \textsc{Mono}}\\{\small \textsc{Trans}}}}
    && Token emb && 83.9 && 67.9 & 62.1 & 63.0 & 64.2 & 51.2 & 61.0 & 64.1 & 52.6 & 51.4 & 50.9 && 61.1 & \\
    &&& \quad + pos emb && \textbf{84.7} && \textbf{73.1} & 65.9 & 66.5 & 66.2 & 16.2 & 59.5 & 65.8 & 51.5 & 56.4 & 19.3 && 56.8 & \\
    &&& \quad + noising && 82.1 && 68.4 & 68.2 & 67.3 & 67.5 & 17.5 & 61.2 & 65.9 & 57.5 & 58.5 & 21.5 && 57.8 & \\
    &&& \quad + adapters && 82.1 && 70.8 & \textbf{70.6} & \underline{\textbf{67.9}} & \textbf{69.1} & \underline{\textbf{61.3}} & \textbf{66.0} & \textbf{67.0} & \textbf{57.5} & \underline{\textbf{60.5}} & \underline{\textbf{61.9}} && \textbf{66.8} & \\
    \bottomrule
  \end{tabular}
\end{small}
\end{center}
\caption{XQuAD results (F1). We bold the best result in each section and underline the overall best result.}
\label{tab:results_xquad_f1}
\end{table*}

We
show $F_1$ scores on XQuAD in Table \ref{tab:results_xquad_f1} (we
include exact match scores in Appendix~\ref{app:results}).
Similar to our findings in the XNLI experiment, 
the vocabulary size has a large impact on 
\jointmulti, and \jointpair 
models with disjoint vocabularies perform the best.
The gap between \monotrans and joint models 
is larger, but \monotrans still performs surprisingly 
well given the nature of the task.
We observe that 
learning language-specific position embeddings 
is helpful in most cases, but completely fails for Turkish and Hindi. 
Interestingly, the exact same pre-trained models 
(after Steps 1 and 2) do obtain competitive 
results in XNLI (\textsection \ref{subsec:xnli}). 
In contrast to results on previous tasks, 
adding adapters to allow a
transferred monolingual model to learn higher
level abstractions in the new language 
significantly improves performance, resulting in a \monotrans model that is comparable
to the best joint system.

\begin{table*}[t]
\begin{center}
\begin{small}
\addtolength{\tabcolsep}{-3.5pt}
  \begin{tabular}{ll lccccccccccccccc}
    \toprule
    && & mono &&& \multicolumn{12}{c}{xx$\rightarrow$en aligned} \\
    \cmidrule{4-5} \cmidrule{7-18}
    && & en &&& en & fr & es & de & el & bg & ru & tr & ar & vi & zh & avg \\
    \midrule
    \multirow{2}{*}{\shortstack{Semantic}}
    && WiC & 59.1 &&& 58.2 & 62.5 & 59.6 & 58.0 & 59.9 & 56.9 & 57.7 & 58.5 & 59.7 & 57.8 & 56.7 & 58.7\\ 
    && SCWS & 45.9	&&& 44.3 & 39.7 & 34.1 & 39.1 & 38.2 & 28.9 & 32.6 & 42.1 & 45.5 & 35.3 & 31.8 & 37.4\\ 
    \midrule 
    \multirow{2}{*}{\shortstack{Syntactic}} 
    && Subject-verb agreement & 86.5 &&& 58.2 & 64.0 & 65.7 & 57.6 & 67.6 & 58.4 & 73.6 & 59.6 & 61.2 & 62.1 & 61.1 & 62.7\\
    && Reflexive anaphora & 79.2 &&& 60.2 & 60.7 & 66.6 & 53.3 & 63.6 & 56.0 & 75.4 & 69.4 & 81.6 & 58.4 & 55.2 & 63.7\\
    \bottomrule
  \end{tabular}
\end{small}
\end{center}
\caption{Semantic and syntactic probing results of a monolingual model and monolingual models transferred to English. Results are on the Word-in-Context (WiC) dev set, the Stanford Contextual Word Similarity (SCWS) test set, and the syntactic evaluation (syn) test set \cite{Marvin2018}. Metrics are accuracy (WiC), Spearman's \emph{r} (SCWS), and macro-averaged accuracy (syn).}
\label{tab:probing}
\end{table*}

\section{Discussion}
\label{sec:discussion}

\paragraph{Joint multilingual training.} We demonstrate 
that sharing subwords across languages is not necessary for mBERT to work, 
contrary to a previous hypothesis by \citet{Pires2019}.
We also do not observe clear improvements by scaling
the joint training to a large number of languages.

Rather than having a joint vs.~disjoint vocabulary or two vs.~multiple languages, 
we find that an important factor 
is the \emph{effective vocabulary size per language}.
When using a joint vocabulary, only a subset of 
the tokens is effectively shared, while the 
rest tends to occur in only one language.
As a result, multiple languages 
compete for allocations in the
shared vocabulary.
We observe that 
multilingual models with larger vocabulary sizes 
obtain consistently better results. 
It is also interesting that our best 
results are generally obtained by the \jointpair systems with a disjoint vocabulary, which guarantees 
that each language is allocated 32k subwords. 
As such, we believe that future work should treat the
effective vocabulary size as an important factor.

\paragraph{Transfer of monolingual representations.} 
\monotrans is competitive even in the most
challenging scenarios.
This indicates that joint multilingual pre-training is not essential for cross-lingual generalization,
suggesting that monolingual models learn linguistic abstractions that generalize across languages. %

To get a better understanding of this phenomenon, we probe the representations of \monotrans. As existing probing datasets are only available in English, we train monolingual representations in non-English languages and transfer them to English.
We probe representations from the resulting
English models with the Word in Context \cite[WiC;][]{wic}, Stanford Contextual Word Similarity \cite[SCWS;][]{scws}, and 
the syntactic evaluation \citep{Marvin2018} datasets.

We provide details
of our experimental setup in Appendix~\ref{app:probing}
and show a summary of our results in Table~\ref{tab:probing}.
The results indicate that 
monolingual semantic representations learned from
non-English languages transfer 
to English to a degree.
On WiC, models transferred from non-English 
languages are comparable with models trained on
English.
On SCWS, while there are more variations, models trained on other
languages still perform surprisingly well.
In contrast, we observe larger gaps in the syntactic evaluation dataset. This suggests that transferring syntactic abstractions is more challenging than semantic abstractions.
We leave a more thorough investigation of
whether joint multilingual pre-training 
reduces to learning a lexical-level alignment for future work.

\paragraph{\clwe.} 
\clwe models---although similar in spirit to \monotrans---are 
only competitive on the easiest and smallest task (MLDoc), 
and perform poorly on the more challenging ones (XNLI and XQuAD). 
While previous work has questioned evaluation methods in this research area \citep{glavas2019properly,artetxe2019bilingual}, our results provide evidence that existing 
methods are 
not competitive in challenging downstream tasks and that mapping between two fixed embedding spaces may be overly restrictive.
For that reason, we think that designing better integration
techniques of \clwe 
to downstream models is an important
future direction.

\paragraph{Lifelong learning.} Humans
learn continuously and accumulate knowledge throughout
their lifetime.
In contrast, existing multilingual models focus
on the scenario where all training data for all languages is available in advance.
The setting to transfer a monolingual model to other 
languages is suitable for the scenario where 
one needs to incorporate new languages into an existing model,
while no longer having access to the original data. 
Such a scenario is of significant practical interest, 
since models are often released 
without the data they are trained on. In that regard, our work provides a baseline for multilingual lifelong learning.

\section{Related Work} \label{sec:related}

\paragraph{Unsupervised lexical multilingual representations.} 
A common approach to learn multilingual representations is based on
cross-lingual word embedding mappings.
These methods learn a set of monolingual word embeddings for each language
and map them to a shared space through a linear transformation. 
Recent approaches perform this mapping with an 
unsupervised initialization based on heuristics \citep{artetxe2018robust} or adversarial training \citep{zhang2017adversarial,conneau2018word}, which is further improved through self-learning \citep{artetxe2017learning}. The same approach has also been adapted for contextual representations \cite{Schuster2019}.

\paragraph{Unsupervised deep multilingual representations.} 
In contrast to the previous approach, which learns a shared multilingual
space at the lexical level,
state-of-the-art methods learn deep representations with a transformer. 
Most of these methods are based on mBERT. 
Extensions to mBERT include scaling it up and incorporating parallel data \cite{lample2019cross},
adding auxiliary pre-training tasks \cite{Huang2019}, 
and encouraging representations of translations to be similar \cite{iclr2020_multilingual}.

Concurrent to this work,  \citet{iclr2020_from_english} propose a more complex approach to transfer a monolingual BERT to other languages 
that achieves results similar to ours. However, 
they find that post-hoc embedding 
learning from a random initialization does not work well.
In contrast, we show that monolingual representations generalize well to other languages and that we can transfer to a new language by learning new subword embeddings.
Contemporaneous work also shows that a shared vocabulary 
is not important for learning multilingual representations \cite{iclr2020_empirical_study,wu2019emerging}, while \citet{Lewis2019mlqa} propose a question answering dataset that is similar in spirit to ours but covers fewer languages and is not parallel across all of them.

\section{Conclusions}
We compared state-of-the-art
multilingual representation learning models
and a monolingual model
that is transferred to new languages
at the lexical level.
We demonstrated that these models perform comparably 
on standard zero-shot cross-lingual transfer benchmarks,
indicating that neither a shared vocabulary nor joint 
pre-training are necessary in multilingual models.
We also showed that a monolingual
model trained on a particular language learns some
semantic abstractions that are generalizable to
other languages in a series of probing experiments.
Our results and analysis contradict previous theories and provide new insights 
into the basis of the generalization abilities of multilingual models.
To provide a more comprehensive benchmark to evaluate cross-lingual models, 
we also released the Cross-lingual Question Answering Dataset (XQuAD).

\section*{Acknowledgements}
We thank Chris Dyer and Phil Blunsom 
for helpful comments on an earlier draft of this paper
and Tyler Liechty for assistance with datasets.

\bibliography{acl2020}
\bibliographystyle{acl_natbib}

\appendix
\begin{table*}[t]
\begin{center}
\begin{small}
\addtolength{\tabcolsep}{-3.5pt}
  \begin{tabular}{lccccccccccc}
    \toprule
    & en & es & de & el & ru & tr & ar & vi & th & zh & hi \\
    \midrule
    Paragraph &
    142.4 & 160.7 & 139.5 & 149.6 & 133.9 & 126.5 & 128.2 & 191.2 & 158.7 & 147.6 & 232.4 \\
    Question &
    11.5 & 13.4 & 11.0 & 11.7 & 10.0 & 9.8 & 10.7 & 14.8 & 11.5 & 10.5 & 18.7 \\
    Answer &
    3.1 & 3.6 & 3.0 & 3.3 & 3.1 & 3.1 & 3.1 & 4.5 & 4.1 & 3.5 & 5.6 \\
    \bottomrule
  \end{tabular}
\end{small}
\end{center}
\caption{Average number of tokens for each language in XQuAD. The statistics were obtained using Jieba for Chinese and the Moses tokenizer for the rest of the languages.}
\label{tab:xquad_stats}
\end{table*}

\begin{table*}[t]
\begin{center}
\begin{small}
\addtolength{\tabcolsep}{-3.5pt}
  \begin{tabular}{cp{7cm}p{7cm}}
    \toprule
    Lang & \multicolumn{1}{c}{Context paragraph w/ answer spans} & \multicolumn{1}{c}{Questions} \\
    \midrule
    en &
    The heat required for boiling the water and supplying the steam can be derived from various sources, most commonly from \textbf{[burning combustible materials]$_1$} with an appropriate supply of air in a closed space (called variously \textbf{[combustion chamber]$_2$}, firebox). In some cases the heat source is a nuclear reactor, geothermal energy, \textbf{[solar]$_3$} energy or waste heat from an internal combustion engine or industrial process. In the case of model or toy steam engines, the heat source can be an \textbf{[electric]$_4$} heating element. &     \vspace{-4\topsep}
    \begin{enumerate}
    \setlength{\itemsep}{3pt}
    \setlength{\parskip}{0pt}
    \item What is the usual source of heat for boiling water in the steam engine?
    \item Aside from firebox, what is another name for the space in which combustible material is burned in the engine?
    \item Along with nuclear, geothermal and internal combustion engine waste heat, what sort of energy might supply the heat for a steam engine?
    \item What type of heating element is often used in toy steam engines?
    \end{enumerate}
    \vspace{-4\topsep}
    \\
    \midrule
    es &
    El calor necesario para hervir el agua y suministrar el vapor puede derivarse de varias fuentes, generalmente de \textbf{[la quema de materiales combustibles]$_1$} con un suministro adecuado de aire en un espacio cerrado (llamado de varias maneras: \textbf{[cámara de combustión]$_2$}, chimenea...). En algunos casos la fuente de calor es un reactor nuclear, energía geotérmica, \textbf{[energía solar]$_3$} o calor residual de un motor de combustión interna o proceso industrial. En el caso de modelos o motores de vapor de juguete, la fuente de calor puede ser un calentador \textbf{[eléctrico]$_4$}. &
    \vspace{-4\topsep}
    \begin{enumerate}
    \setlength{\itemsep}{3pt}
    \setlength{\parskip}{0pt}
    \item ¿Cuál es la fuente de calor habitual para hacer hervir el agua en la máquina de vapor?
    \item Aparte de cámara de combustión, ¿qué otro nombre que se le da al espacio en el que se quema el material combustible en el motor?
    \item Junto con el calor residual de la energía nuclear, geotérmica y de los motores de combustión interna, ¿qué tipo de energía podría suministrar el calor para una máquina de vapor?
    \item ¿Qué tipo de elemento calefactor se utiliza a menudo en las máquinas de vapor de juguete?
    \end{enumerate}
    \vspace{-4\topsep}
    \\
    \midrule
    zh &
    \begin{CJK}{UTF8}{gbsn}
    让水沸腾以提供蒸汽所需热量有多种来源，最常见的是在封闭空间（别称有 \textbf{[燃烧室]$_2$} 、火箱）中供应适量空气来 \textbf{[燃烧可燃材料]$_1$} 。在某些情况下，热源是核反应堆、地热能、 \textbf{[太阳能]$_3$} 或来自内燃机或工业过程的废气。如果是模型或玩具蒸汽发动机，还可以将 \textbf{[电]$_4$} 加热元件作为热源。
    \end{CJK} &
    \vspace{-4\topsep}
    \begin{CJK}{UTF8}{gbsn}
    \begin{enumerate}
    \setlength{\itemsep}{3pt}
    \setlength{\parskip}{0pt}
    \item 蒸汽机中让水沸腾的常用热源是什么?
    \item 除了火箱之外，发动机内燃烧可燃材料的空间的别名是什么?
    \item 除了核能、地热能和内燃机废气以外，还有什么热源可以为蒸汽机供能?
    \item 玩具蒸汽机通常使用什么类型的加热元件?
    \end{enumerate}
    \end{CJK}
    \vspace{-4\topsep}
    \\
     \bottomrule
  \end{tabular}
\end{small}
\end{center}
\caption{An example from XQuAD. The full dataset consists of 240 such parallel instances in 11 languages.}
\label{tab:xquad_example}
\end{table*}

\section{Training details} \label{app:hyperparameters}

In contrast to \citet{you2019reducing}, we train with a sequence length of 512 from the beginning, instead of dividing training into two stages. For our proposed approach, we pre-train a single English model for 250k steps, and perform another 250k steps to transfer it to every other language.

For the fine-tuning, we use Adam with a learning rate of 2e-5, a batch size of 32, and train for 2 epochs. The rest of the hyperparameters follow \citet{devlin2018bert}. For adapters, we follow the hyperparameters employed by \citet{Houlsby2019}. For our proposed model using noised fine-tuning, we set the standard deviation of the Gaussian noise to 0.075 and the mean to 0.

\section{XQuAD dataset details} \label{app:xquad_dataset}

XQuAD consists of a subset of 240 context paragraphs and 1190 question-answer pairs 
from the development set of SQuAD v1.1 \citep{squad} together with their translations into 
10 other languages: Spanish, German, Greek, Russian, Turkish, Arabic, Vietnamese, Thai, Chinese, and Hindi. Table \ref{tab:xquad_stats} comprises some statistics of the dataset, while Table \ref{tab:xquad_example} shows one example from it.

So as to guarantee the diversity of the dataset, we selected 5 context paragraphs at random from each of the 48 documents in the SQuAD 1.1 development set, and translate both the context paragraphs themselves as well as all their corresponding questions. The translations were done by professional human translators through the Gengo\footnote{\url{https://gengo.com}} service. The translation workload was divided into 10 batches for each language, which were submitted separately to Gengo. As a consequence, different parts of the dataset might have been translated by different translators. However, we did guarantee that all paragraphs and questions from the same document were submitted in the same batch to make sure that their translations were consistent. Translators were specifically instructed to transliterate all named entities to the target language following the same conventions used in Wikipedia, from which the English context paragraphs in SQuAD originally come.

In order to facilitate easy annotations of answer spans, we chose the most frequent answer for each question and marked its beginning and end in the context paragraph through placeholder symbols (e.g. \textit{``this is *0* an example span \#0\# delimited by placeholders''}). Translators were instructed to keep the placeholders in the relevant position in their translations, and had access to an online validator to automatically verify that the format of their output was correct.

\section{Additional results} \label{app:results}

We show the complete results for cross-lingual word embedding mappings and joint multilingual training on MLDoc and PAWS-X in Table \ref{tab:results_mldoc_paws_clwe_joint_multi}. Table \ref{tab:results_xquad_em} reports exact match results on XQuAD, while Table \ref{tab:results_xquad_f1_clwe_join_multi} reports results for all cross-lingual word embedding mappings and joint multilingual training variants.

\begin{table*}[t]
\begin{center}
\begin{small}
\addtolength{\tabcolsep}{-3.5pt}
  \begin{tabular}{rcrlrcrcccccrcrrcrccccrcr}
    \toprule
    &&&&& \multicolumn{9}{c}{MLDoc} &&& \multicolumn{8}{c}{PAWS-X} & \\
    \cmidrule{6-14} \cmidrule{17-24}
    & & & & & en & & fr & es & de & ru & zh && avg &&& en && fr & es & de & zh && avg & \\
    \midrule
    & \multirow{4}{*}{\shortstack{\clwe}}
    && 300d ident && 93.1 && 85.2 & 74.8 & 86.5 & 67.4 & 72.7 && 79.9 &&& 92.8 && 83.9 & 84.7 & 81.1 & 72.9 && 83.1 & \\
    &&& 300d unsup && 93.1 && 85.0 & 75.0 & 86.1 & 68.8 & 76.0 && 80.7 &&& 92.8 && 83.9 & 84.2 & 81.3 & 73.5 && 83.1 & \\
    &&& 768d ident && 94.7 && 87.3 & 77.0 & 88.7 & 67.6 & 78.3 && 82.3 &&& 92.8 && 85.2 & 85.5 & 81.6 & 72.5 && 83.5 & \\
    &&& 768d unsup && 94.7 && 87.5 & 76.9 & 88.1 & 67.6 & 72.7 && 81.2 &&& 92.8 && 84.3 & 85.5 & 81.8 & 72.1 && 83.3 & \\
    \midrule
    & \multirow{4}{*}{\shortstack{{\small \textsc{Joint}}\\{\small \textsc{Multi}}}}
    && 32k voc && 92.6 && 81.7 & 75.8 & 85.4 & 71.5 & 66.6 && 78.9 &&& 91.9 && 83.8 & 83.3 & 82.6 & 75.8 && 83.5 & \\
    &&& 64k voc && 92.8 && 80.8 & 75.9 & 84.4 & 67.4 & 64.8 && 77.7 &&& 93.7 && 86.9 & 87.8 & 85.8 & 80.1 && 86.8 & \\
    &&& 100k voc && 92.2 && 74.0 & 77.2 & 86.1 & 66.8 & 63.8 && 76.7 &&& 93.1 && 85.9 & 86.5 & 84.1 & 76.3 && 85.2 & \\
    &&& 200k voc && 91.9 && 82.1 & 80.9 & 89.3 & 71.8 & 66.2 && 80.4 &&& 93.8 && 87.7 & 87.5 & 87.3 & 78.8 && 87.0 & \\
    \bottomrule
  \end{tabular}
\end{small}
\end{center}
\caption{MLDoc and PAWS-X results (accuracy) for all \clwe and \jointmulti variants.}
\label{tab:results_mldoc_paws_clwe_joint_multi}
\end{table*}

\begin{table*}[t]
\begin{center}
\begin{small}
\addtolength{\tabcolsep}{-3.5pt}
  \begin{tabular}{rcrlrcrccccccccccrcr}
    \toprule
    & & & & & en && es & de & el & ru & tr & ar & vi & th & zh & hi && avg & \\
    \midrule
    & \multirow{4}{*}{\shortstack{\clwe}}
    && 300d ident && 72.5 && 39.7 & 33.6 & 23.5 & 29.9 & 11.8 & 18.5 & 16.1 & 16.5 & 17.9 & 10.0 && 26.4 & \\
    &&& 300d unsup && 72.5 && 39.2 & 34.5 & 24.8 & 30.4 & 12.2 & 14.7 & 6.5 & 16.0 & 16.1 & 10.4 && 25.2 & \\
    &&& 768d ident && 73.1 && 40.6 & 32.9 & 20.1 & 30.7 & 10.8 & 14.2 & 11.8 & 12.3 & 14.0 & 9.1 && 24.5 & \\
    &&& 768d unsup && 73.1 && 41.5 & 31.8 & 21.0 & 31.0 & 12.1 & 14.1 & 10.5 & 10.0 & 13.2 & 10.2 && 24.4 & \\
    \midrule
    & \multirow{4}{*}{\shortstack{{\small \textsc{Joint}}\\{\small \textsc{Multi}}}}
    && 32k voc && 68.3 && 41.3 & 44.3 & 31.8 & 45.0 & 28.5 & 36.2 & 36.9 & 39.2 & 40.1 & 27.5 && 39.9 & \\
    &&& 64k voc && 71.3 && 48.2 & 49.9 & 40.2 & 50.9 & 33.7 & 41.5 & 45.0 & 43.7 & 36.9 & 36.8 && 45.3 & \\
    &&& 100k voc && 71.5 && 49.8 & 51.2 & 41.1 & 51.8 & 33.0 & 43.7 & 45.3 & 44.5 & 40.8 & 36.6 && 46.3 & \\
    &&& 200k voc && 72.1 && 55.3 & 55.2 & 48.0 & 52.7 & 40.1 & 46.6 & 47.6 & 45.8 & 38.5 & 42.3 && 49.5 & \\
    \midrule
    & \multirow{2}{*}{\shortstack{{\small \textsc{Joint}}\\{\small \textsc{Pair}}}}
    && Joint voc && 71.7 && 47.8 & 57.6 & 38.2 & 53.4 & 35.0 & 47.4 & 49.7 & 44.3 & 47.1 & 38.8 && 48.3 & \\
    &&& Disjoint voc && 72.2 && 52.5 & 56.5 & 47.8 & 55.0 & 43.7 & 49.0 & 49.2 & 43.9 & 50.0 & 39.1 && 50.8 & \\
    \midrule
    & \multirow{4}{*}{\shortstack{{\small \textsc{Mono}}\\{\small \textsc{Trans}}}}
    && Subword emb && 72.3 && 47.4 & 42.4 & 43.3 & 46.4 & 30.1 & 42.6 & 45.1 & 39.0 & 39.0 & 32.4 && 43.6 & \\
    &&& \quad + pos emb && 72.9 && 54.3 & 48.4 & 47.3 & 47.6 & 6.1 & 41.1 & 47.6 & 38.6 & 45.0 & 9.0 && 41.6 & \\
    &&& \quad + noising && 69.6 && 51.2 & 52.4 & 50.2 & 51.0 & 6.9 & 43.0 & 46.3 & 46.4 & 48.1 & 10.7 && 43.2 & \\
    &&& \quad + adapters && 69.6 && 51.4 & 51.4 & 50.2 & 51.4 & 44.5 & 48.8 & 47.7 & 45.6 & 49.2 & 45.1 && 50.5 & \\
    \bottomrule
  \end{tabular}
\end{small}
\end{center}
\caption{XQuAD results (exact match).}
\label{tab:results_xquad_em}
\end{table*}

\begin{table*}[t]
\begin{center}
\begin{small}
\addtolength{\tabcolsep}{-3.5pt}
  \begin{tabular}{rcrlrcrccccccccccrcr}
    \toprule
    & & & & & en && es & de & el & ru & tr & ar & vi & th & zh & hi && avg & \\
    \midrule
    & \multirow{4}{*}{\shortstack{\clwe}}
    && 300d ident && 84.1 && 56.8 & 51.3 & 43.4 & 47.4 & 25.5 & 35.5 & 34.5 & 28.7 & 25.3 & 22.1 && 41.3 & \\
    &&& 300d unsup && 84.1 && 56.8 & 51.8 & 42.7 & 48.5 & 24.4 & 31.5 & 20.5 & 29.8 & 26.6 & 23.1 && 40.0 & \\
    &&& 768d ident && 84.2 && 58.0 & 51.2 & 41.1 & 48.3 & 24.2 & 32.8 & 29.7 & 23.8 & 19.9 & 21.7 && 39.5 & \\
    &&& 768d unsup && 84.2 && 58.9 & 50.3 & 41.0 & 48.5 & 25.8 & 31.3 & 27.3 & 24.4 & 20.9 & 21.6 && 39.5 & \\
    \midrule
    & \multirow{4}{*}{\shortstack{{\small \textsc{Joint}}\\{\small \textsc{Multi}}}}
    && 32k voc && 79.3 && 59.5 & 60.3 & 49.6 & 59.7 & 42.9 & 52.3 & 53.6 & 49.3 & 50.2 & 42.3 && 54.5 & \\
    &&& 64k voc && 82.3 && 66.5 & 67.1 & 60.9 & 67.0 & 50.3 & 59.4 & 62.9 & 55.1 & 49.2 & 52.2 && 61.2 & \\
    &&& 100k voc && 82.6 && 68.9 & 68.9 & 61.0 & 67.8 & 48.1 & 62.1 & 65.6 & 57.0 & 52.3 & 53.5 && 62.5 & \\
    &&& 200k voc && 82.7 && 74.3 & 71.3 & 67.1 & 70.2 & 56.6 & 64.8 & 67.6 & 58.6 & 51.5 & 58.3 && 65.7 & \\
    \bottomrule
  \end{tabular}
\end{small}
\end{center}
\caption{XQuAD results (F1) for all \clwe and \jointmulti variants.}
\label{tab:results_xquad_f1_clwe_join_multi}
\end{table*}

\section{Probing experiments} \label{app:probing}

As probing tasks are only available in English, we train monolingual models in each $L_2$ of XNLI and then align them to English. To control for the amount of data, we use 3M sentences both for pre-training and alignment in every language.\footnote{We leave out Thai, Hindi, Swahili, and Urdu as their corpus size is smaller than 3M.}

\paragraph{Semantic probing} We evaluate the representations on two semantic probing tasks, the Word in Context \cite[WiC;][]{wic} and Stanford Contextual Word Similarity \cite[SCWS;][]{scws} datasets. WiC is a binary classification task, which requires the model to determine if the occurrences of a word in two contexts refer to the same or different meanings. SCWS requires estimating the semantic similarity of word pairs that occur in context. For WiC, we train a linear classifier on top of the fixed sentence pair representation. For SCWS, we obtain the contextual representations of the target word in each sentence by averaging its constituent word pieces, and calculate their cosine similarity.

\paragraph{Syntactic probing} We evaluate the same models in the syntactic probing dataset of \citet{Marvin2018} following the same setup as \citet{Goldberg2019}. Given minimally different pairs of English sentences, the task is to identify which of them is grammatical. Following \citet{Goldberg2019}, we feed each sentence into the model masking the word in which it differs from its pair, and pick the one to which the masked language model assigns the highest probability mass. Similar to \citet{Goldberg2019}, we discard all sentence pairs from the \citet{Marvin2018} dataset that differ in more than one subword token. Table \ref{tab:syntactic_probing_stats} reports the resulting coverage split into different categories, and we show the full results in Table \ref{tab:full_syntactic_probing}.

\begin{table*}[t]
\begin{center}
\begin{small}
\addtolength{\tabcolsep}{-3.5pt}
  \begin{tabular}{lccc}
    \toprule
    & coverage \\
    \midrule
    \textbf{Subject-verb agreement} \\
    \midrule
    Simple & 80 / 140 (57.1\%) \\
    In a sentential complement & 960 / 1680 (57.1\%) \\
    Short VP coordination & 480 / 840 (57.1\%) \\
    Long VP coordination & 320 / 400 (80.0\%) \\
    Across a prepositional phrase & 15200 / 22400 (67.9\%) \\
    Across a subject relative clause & 6400 / 11200 (57.1\%) \\
    Across an object relative clause & 17600 / 22400 (78.6\%) \\
    Across an object relative (no that) & 17600 / 22400 (78.6\%) \\
    In an object relative clause & 5600 / 22400 (25.0\%) \\
    In an object relative (no that) & 5600 / 22400 (25.0\%) \\
    \midrule
    \textbf{Reflexive anaphora} \\
    \midrule
    Simple & 280 / 280 (100.0\%) \\
    In a sentential complement & 3360 / 3360 (100.0\%) \\
    Across a relative clause & 22400 / 22400 (100.0\%) \\
    \bottomrule
  \end{tabular}
\end{small}
\end{center}
\caption{Coverage of our systems for the syntactic probing dataset. We report the number of pairs in the original dataset by \citet{Marvin2018}, those covered by the vocabulary of our systems and thus used in our experiments, and the corresponding percentage.}
\label{tab:syntactic_probing_stats}
\end{table*}

\begin{table*}[t]
\begin{center}
\begin{small}
\addtolength{\tabcolsep}{-3.5pt}
  \begin{tabular}{lccccccccccccccc}
    \toprule
    & mono &&& \multicolumn{12}{c}{xx$\rightarrow$en aligned} \\
    \cmidrule{2-3} \cmidrule{5-16}
    & en &&& en & fr & es & de & el & bg & ru & tr & ar & vi & zh & avg \\
    \midrule
    \textbf{Subject-verb agreement} \\
    \midrule
    Simple & 91.2 &&& 76.2 & 90.0 & 93.8 & 56.2 & 97.5 & 56.2 & 78.8 & 72.5 & 67.5 & 81.2 & 71.2 & 76.5 \\
    In a sentential complement & 99.0 &&& 65.7 & 94.0 & 92.1 & 62.7 & 98.3 & 80.7 & 74.1 & 89.7 & 71.5 & 78.9 & 79.6 & 80.7 \\
    Short VP coordination & 100.0 &&& 64.8 & 66.9 & 69.8 & 64.4 & 77.9 & 60.2 & 88.8 & 76.7 & 73.3 & 62.7 & 64.4 & 70.0 \\
    Long VP coordination & 96.2 &&& 58.8 & 53.4 & 60.0 & 67.5 & 62.5 & 59.4 & 92.8 & 62.8 & 75.3 & 62.5 & 64.4 & 65.4 \\
    Across a prepositional phrase & 89.7 &&& 56.9 & 54.6 & 52.8 & 53.4 & 53.4 & 54.6 & 79.6 & 54.3 & 59.9 & 57.9 & 56.5 & 57.6 \\
    Across a subject relative clause & 91.6 &&& 49.9 & 51.9 & 48.3 & 52.0 & 53.2 & 56.2 & 78.1 & 48.6 & 58.9 & 55.4 & 52.3 & 55.0 \\
    Across an object relative clause & 79.2 &&& 52.9 & 56.2 & 53.3 & 52.4 & 56.6 & 57.0 & 63.1 & 52.3 & 59.0 & 54.9 & 54.5 & 55.7 \\
    Across an object relative (no that) & 77.1 &&& 54.1 & 55.9 & 55.9 & 53.1 & 56.2 & 59.7 & 63.3 & 53.1 & 54.9 & 55.9 & 56.8 & 56.3 \\
    In an object relative clause & 74.6 &&& 50.6 & 59.9 & 66.4 & 59.4 & 61.1 & 49.8 & 60.4 & 42.6 & 45.3 & 56.9 & 56.3 & 55.3 \\
    In an object relative (no that) & 66.6 &&& 51.7 & 57.1 & 64.9 & 54.9 & 59.4 & 49.9 & 57.0 & 43.7 & 46.6 & 54.9 & 55.4 & 54.1 \\
    \textit{Macro-average} & 86.5 &&& 58.2 & 64.0 & 65.7 & 57.6 & 67.6 & 58.4 & 73.6 & 59.6 & 61.2 & 62.1 & 61.1 & 62.7\\
    \midrule
    \textbf{Reflexive anaphora} \\
    \midrule
    Simple & 90.0 &&& 69.3 & 63.6 & 67.9 & 55.0 & 69.3 & 56.4 & 89.3 & 75.0 & 87.1 & 58.6 & 60.7 & 68.4 \\
    In a sentential complement & 82.0 &&& 56.3 & 63.9 & 73.2 & 52.7 & 65.7 & 59.1 & 70.8 & 71.7 & 84.5 & 59.8 & 53.9 & 64.7 \\
    Across a relative clause & 65.6 &&& 55.0 & 54.5 & 58.6 & 52.3 & 55.8 & 52.5 & 66.1 & 61.4 & 73.3 & 56.9 & 50.9 & 57.9 \\
    \textit{Macro-average} & 79.2 &&& 60.2 & 60.7 & 66.6 & 53.3 & 63.6 & 56.0 & 75.4 & 69.4 & 81.6 & 58.4 & 55.2 & 63.7\\
    \bottomrule
  \end{tabular}
\end{small}
\end{center}
\caption{Complete syntactic probing results (accuracy) of a monolingual model and monolingual models transferred to English on the syntactic evaluation test set \cite{Marvin2018}.}
\label{tab:full_syntactic_probing}
\end{table*}

\end{document}